\documentclass{article}
\usepackage{spconf,amsmath,amssymb,graphicx}
\usepackage{bm}
\usepackage{multirow}
\usepackage{booktabs}
\usepackage{makecell}
\usepackage{float}
\usepackage{url}
\usepackage{hyperref}
\usepackage{caption}
\usepackage{comment}
\usepackage{color}

\newcommand{\ie}{\textit{i.e.}}

\title{Generalizable Face Forgery Detection via Separable Prompt Learning}

\name{Enrui Yang$^1$, Baoyuan Wu$^2$, Yuezun Li$^{1,*}$\thanks{* Corresponding author.}}
\address{$^1$ School of Computer Science and Technology, Ocean University of China, Qingdao, China \\
$^2$ School of Artificial Intelligence, The Chinese University of Hong Kong, Shenzhen, China}

\begin{document}
%
\maketitle
\begin{abstract}
Detecting face forgeries using CLIP has recently emerged as a promising direction. However, most existing methods focus on adapting its visual encoder, leaving the potential of the textual encoder largely underexplored. In this paper, we propose Separable Prompt Learning (\textbf{SePL}) to better exploit the text modality, which further enhances the detection capacity. Specifically, SePL distills the forgery knowledge from CLIP via two separate learnable prompts, supported by a cross-modality alignment strategy and dedicated objectives. Extensive experiments demonstrate that our method achieves superior performance under both cross-dataset and cross-method evaluation. The code has been released at \url{https://github.com/OUC-YER/SePL-DeepfakeDetection}.
\end{abstract}
\begin{keywords}
Face forgery detection, deepfake detection, CLIP, prompt learning
\end{keywords}
\section{Introduction}
\label{sec:intro}

The rapid advancement of generative models has significantly increased the realism of deepfakes, making them increasingly indistinguishable to the human eye. The misuse of this technique enables large-scale fabrication, posing serious societal concerns~\cite{deepfake_survey}. 
In response, substantial efforts~\cite{udd,svd,recce,sbi,forensics_adapter} have been devoted to face forgery detection.

Since deepfake techniques continue to evolve, a critical challenge for real-world deployment is the detection of unseen forgery types, \ie, generalizability. Several methods have been proposed to capture generic forgery patterns, either by training on carefully designed pseudo-fake samples~\cite{sbi,sladd} or by adopting feature disentanglement strategies~\cite{ucf}.
However, as they are trained on known data, their ability to generalize is still limited.


CLIP is a recently prevalent vision--language model that has demonstrated strong effectiveness in a wide range of visual tasks~\cite{clip}. Pretrained on large-scale image--text pairs, it possesses versatile prior knowledge for visual content. Motivated by this capability, several recent studies leverage its visual encoder to improve generalizability through adapter modules~\cite{forensics_adapter} or parameter-efficient fine-tuning (PEFT) strategies~\cite{svd}. Nevertheless, the potential of the textual encoder is largely underexplored. While several methods~\cite{repdfd,c2p_clip,deepfakeclip} incorporate the text modality, they largely follow the strategies developed for general vision tasks~\cite{coop}, using either manually predefined prompts or simple learnable prefixes. Since forgery traces are considerably more subtle than semantic visual features, such designs may not fully unleash the potential of textual modality.

\begin{figure}[t]
\centering
\includegraphics[width=\columnwidth]{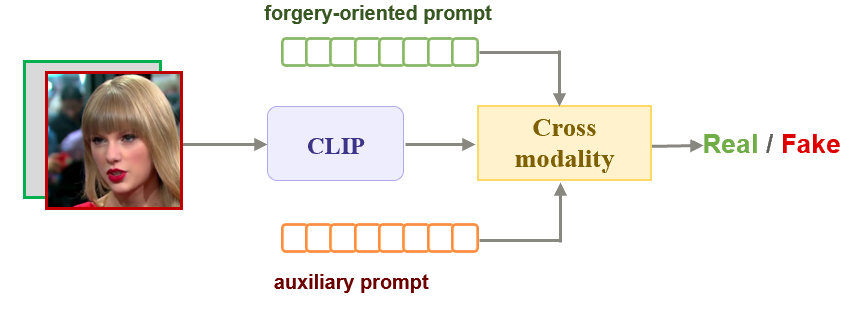} 
\vspace{-0.7cm}
\caption{Overview of the proposed SePL framework. It leverages two separate prompts for CLIP-based deepfake detection.}
\label{fig:pipeline0}
\end{figure}

In this paper, we propose Separable Prompt Learning (\textbf{SePL}) to strengthen the instructive role in face forgery detection. As shown in Fig.~\ref{fig:pipeline0}, SePL first designs two types of learnable prompts with distinct learning roles: a \textit{forgery-oriented prompt} that instructs the capture of condensed forgery cues, and an \textit{auxiliary prompt} that regularizes its learning. A cross-modality alignment strategy is then introduced to distill forgery knowledge from CLIP visual features using the forgery-oriented prompt. 
To make this process work as expected, we design a set of dedicated objectives to guide the two prompts toward their intended roles. Through this simple design, our method further unlocks CLIP capacity and achieves superior generalization performance across diverse evaluation settings. Extensive experiments on widely used benchmarks demonstrate our effectiveness, highlighting the value of the textual modality for generalizable face forgery detection.

\begin{figure*}[t]
\centering
\includegraphics[width=\textwidth]{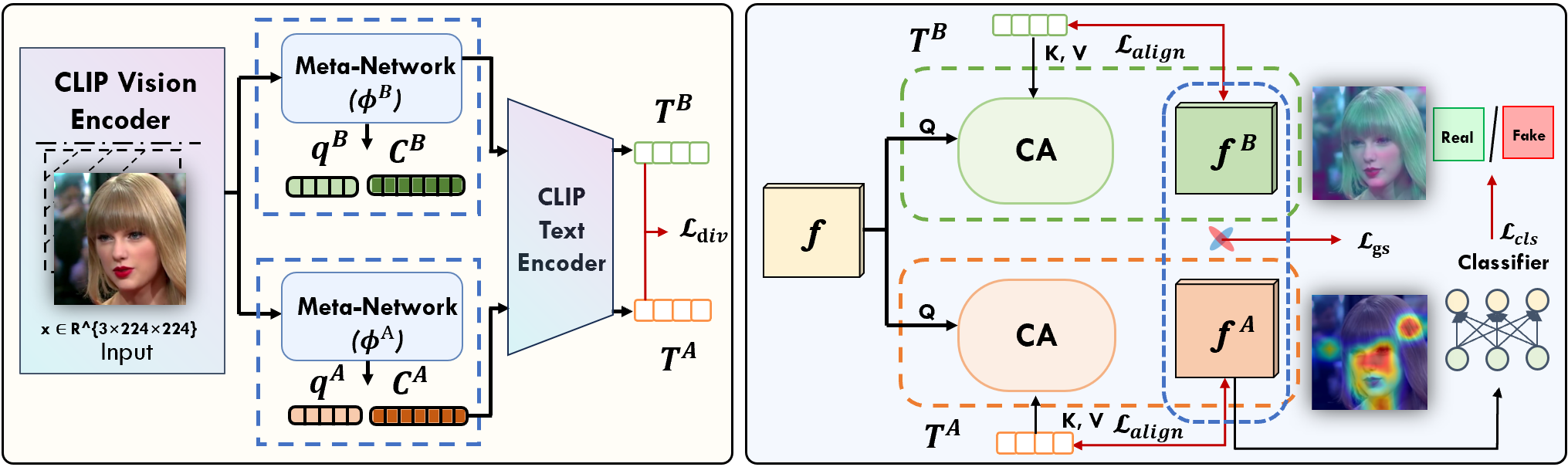}
\vspace{-0.6cm}
\caption{Pipeline of SePL. (left) Instance-level conditional context vectors $\bm{q}^{\mathcal{A}}, \bm{q}^{\mathcal{B}}$ produced by meta-networks are combined with global base context vectors $\bm{C}^{\mathcal{A}}, \bm{C}^{\mathcal{B}}$ to form forgery-oriented and auxiliary prompts. (right) The corresponding embeddings guide the distillation of forgery knowledge $\bm{f}^{\mathcal{A}}$ from CLIP visual features $\bm{f}$ via cross-modality alignment.}
\label{fig:pipeline1}
\end{figure*}

\section{Method}
\label{sec:method}

We describe a Separable Prompt Learning (SePL) framework to enhance CLIP-based face forgery detection. In the following, we first introduce the design of separable prompts (Sec.~\ref{ssec:prompt_design}), and then present the cross-modality alignment strategy (Sec.~\ref{ssec:cma}). Last, we describe the training setup that enables separable prompt learning (Sec.~\ref{ssec:train_test}).

\subsection{Separable Prompt Design}
\label{ssec:prompt_design}

We construct a forgery-oriented prompt and an auxiliary prompt, denoted as $\bm{P}^{\mathcal{A}}$ and $\bm{P}^{\mathcal{B}}$, which are formulated as
\begin{equation}
    \bm{P}^{\mathcal{A}} = [\,\bm{q}^{\mathcal{A}}, \bm{C}^{\mathcal{A}} \,], \quad \bm{P}^{\mathcal{B}} = [\,\bm{q}^{\mathcal{B}}, \bm{C}^{\mathcal{B}} \,].
\end{equation}
$\bm{P}^{\mathcal{A}}$ consists of an instance-level conditional context vector $\bm{q}^{\mathcal{A}}$ and a global base context vector $\bm{C}^{\mathcal{A}}$. Denote the visual encoder as $\bm{h}$ and an input image as $\bm{x}$. To obtain $\bm{q}^{\mathcal{A}}$, we design a lightweight meta-network $\bm{\phi}^{\mathcal{A}}$, which maps the visual feature $\bm{f} = \bm{h}(\bm{x})$ to $\bm{q}^{\mathcal{A}} = \bm{\phi}^{\mathcal{A}}(\bm{f})$. The global base context $\bm{C}^{\mathcal{A}}$ comprises $k$ learnable vectors, \ie, $\bm{C}^{\mathcal{A}} = [ \bm{c}^{\mathcal{A}}_1, \ldots, \bm{c}^{\mathcal{A}}_k ]$, shared across all samples. 
The formulation of $\bm{P}^{\mathcal{B}}$ follows the same design, with $\bm{q}^{\mathcal{B}}$ produced by another meta-network $\bm{\phi}^{\mathcal{B}}$ and a separate global context $\bm{C}^{\mathcal{B}} = [ \bm{c}^{\mathcal{B}}_1, \ldots, \bm{c}^{\mathcal{B}}_k ]$. These prompts are then converted to tokens using the textual encoder $\bm{g}$ as $\bm{T}^{\mathcal{A}} = \bm{g}(\bm{P}^{\mathcal{A}})$ and $\bm{T}^{\mathcal{B}} = \bm{g}(\bm{P}^{\mathcal{B}})$.

This part only describes the structure of these prompts. In the following two sections, we describe how they are learned as expected and how to distill forgery knowledge from CLIP. 

\subsection{Cross-modality Alignment}
\label{ssec:cma}

This strategy is to distill forgery knowledge using forgery-oriented prompt. The process is defined as
\begin{equation}
\begin{array}{c}
    Q = \bm{f}W_Q, \;\; K=\bm{T}^{\mathcal{A}}W_K, \;\; V=\bm{T}^{\mathcal{A}}W_V, \\[0.1cm]
    \bm{f}^{\mathcal{A}} = \mathrm{FFN}\left (\mathrm{LN}(\bm{f} + \mathrm{CA}(Q,K,V) ) \right ),
\end{array}
\end{equation}
where the visual feature $\bm{f}$ serves as the query, while $\bm{T}^{\mathcal{A}}$ provides both keys and values. Here, $W_Q,W_K,W_V$ are projection matrices. The cross-attention(CA) output is added to the original feature via a residual connection, followed by layer normalization (LN) and a feed-forward network (FFN), yielding the aligned feature $\bm{f}^{\mathcal{A}}$, which is used for final detection. 

The same procedure is applied with $\bm{T}^{\mathcal{B}}$ to obtain $\bm{f}^{\mathcal{B}}$, which is used to assist the learning of forgery-oriented prompt. The overall pipeline is illustrated in Fig.~\ref{fig:pipeline1}.

\begin{table*}[!t]
\centering
\caption{Comparison with state-of-the-art deepfake detection methods on cross-dataset and cross-method evaluations. }
\vspace{-0.3cm}
\label{tab:cross_dataset}
\resizebox{\textwidth}{!}{
\begin{tabular}{l ccccc c | cccccccc c}
\toprule
\multirow{2}{*}{Methods}
& \multicolumn{5}{c}{Cross-dataset Evaluation}
& \multirow{2}{*}{Avg.}
& \multicolumn{8}{c}{Cross-method Evaluation}
& \multirow{2}{*}{Avg.} \\
\cmidrule(lr){2-6} \cmidrule(lr){8-15}
 & CDF-v2 & DFD & DFDC & DFDCP & WDF &
 & UniFace & BleFace & MobSwap & e4s & FaceDan & FSGAN & InSwap & SimSwap & \\
\midrule
F3Net~\cite{f3net}   & 0.789 & 0.844 & 0.718 & 0.749 & 0.728 & 0.766 & 0.809 & 0.808 & 0.867 & 0.494 & 0.717 & 0.845 & 0.757 & 0.674 & 0.746 \\
SPSL~\cite{spsl}    & 0.799 & 0.871 & 0.724 & 0.770 & 0.702 & 0.773 & 0.747 & 0.748 & 0.885 & 0.514 & 0.666 & 0.812 & 0.643 & 0.665 & 0.710 \\
SRM~\cite{srm}       & 0.840 & 0.885 & 0.695 & 0.728 & 0.702 & 0.770 & 0.749 & 0.704 & 0.779 & 0.704 & 0.659 & 0.772 & 0.793 & 0.694 & 0.732 \\
CORE~\cite{core}     & 0.809 & 0.882 & 0.721 & 0.720 & 0.724 & 0.771 & 0.871 & 0.843 & 0.959 & 0.679 & 0.774 & 0.958 & 0.855 & 0.724 & 0.833 \\
RECCE~\cite{recce}   & 0.823 & 0.891 & 0.696 & 0.734 & 0.756 & 0.780 & 0.898 & 0.832 & 0.925 & 0.683 & 0.848 & 0.949 & 0.848 & 0.768 & 0.844 \\
SLADD~\cite{sladd}   & 0.837 & 0.904 & 0.772 & 0.756 & 0.690 & 0.792 & 0.878 & 0.882 & 0.954 & 0.765 & 0.825 & 0.943 & 0.879 & 0.794 & 0.865 \\
SBI~\cite{sbi}       & 0.886 & 0.827 & 0.717 & 0.848 & 0.703 & 0.796 & 0.724 & 0.891 & 0.952 & 0.750 & 0.594 & 0.803 & 0.712 & 0.701 & 0.766 \\
UCF~\cite{ucf}       & 0.837 & 0.867 & 0.742 & 0.770 & 0.774 & 0.798 & 0.831 & 0.827 & 0.950 & 0.731 & 0.862 & 0.937 & 0.809 & 0.647 & 0.824 \\
IID~\cite{iid}       & 0.838 & 0.939 & 0.700 & 0.689 & 0.666 & 0.766 & 0.839 & 0.789 & 0.888 & 0.766 & 0.844 & 0.927 & 0.789 & 0.644 & 0.811 \\
LSDA~\cite{lsda}    & 0.875 & 0.881 & 0.701 & 0.812 & 0.797 & 0.813 & 0.872 & 0.875 & 0.930 & 0.694 & 0.721 & 0.939 & 0.855 & 0.793 & 0.835 \\
ProDet~\cite{prodet} & 0.926 & 0.901 & 0.707 & 0.828 & 0.781 & 0.829 & 0.908 & 0.929 & 0.975 & 0.771 & 0.747 & 0.928 & 0.837 & 0.844 & 0.867 \\
CDFA~\cite{cdfa}     & 0.938 & 0.954 & 0.830 & 0.881 & 0.796 & 0.880 & 0.762 & 0.756 & 0.823 & 0.631 & 0.803 & 0.942 & 0.772 & 0.757 & 0.781 \\
Hybrid~\cite{hybrid}  & 0.945 & 0.967 & 0.796 & 0.925 & -     & -     & -     & -     & -     & -     & -     & -     & -     & -     & -     \\
RepDFD~\cite{repdfd}  & 0.899 & -     & 0.809 & \textbf{0.950} & - & - & - & - & - & - & - & - & - & - & - \\
UDD~\cite{udd}        & 0.931 & -     & 0.812 & 0.880 & -     & -     & -     & -     & -     & -     & -     & -     & -     & -     & -     \\
Effort~\cite{svd}    & 0.956 & 0.965 & 0.843 & 0.909 & \textbf{0.848} & 0.904 & 0.962 & 0.873 & 0.953 & \textbf{0.983} & 0.926 & 0.957 & 0.936 & 0.926 & 0.940 \\
ForAda~\cite{forensics_adapter}    & 0.957 & \textbf{0.972} & \textbf{0.872} & 0.929 & 0.805 & 0.907 & 0.954 & 0.888 & 0.981 & 0.971 & 0.950 & \textbf{0.977} & 0.952 & 0.920 & 0.949 \\
\midrule
\textbf{SePL (Ours)}
& \textbf{0.960} & \textbf{0.972} & 0.866 & 0.905 & 0.838 & \textbf{0.908}
& \textbf{0.981} & \textbf{0.960} & \textbf{0.985} & \textbf{0.983} & \textbf{0.974} & \textbf{0.977} & \textbf{0.978} & \textbf{0.973} & \textbf{0.976} \\
\bottomrule
\end{tabular}}
\end{table*}

\subsection{Training}
\label{ssec:train_test}

The key challenge is how to make the forgery-oriented prompt work as expected, \ie, distilling forgery knowledge from CLIP. To achieve this, we design a two-stage training strategy with a set of objectives.

\smallskip
\noindent\textbf{Pretraining.}
Since precise supervision for these prompts is unavailable, we introduce a pretraining strategy to guide their learning implicitly. Motivated by the distinction between semantic and forgery-related information~\cite{svd}, we first align the auxiliary prompt with CLIP to capture semantic information through an instance-level contrastive objective. During the second stage, the auxiliary branch provides a reference for regularizing the forgery-oriented branch
by penalizing the similarity between their features. 
Note that this process encourages the forgery-oriented branch to focus on forgery cues, \textit{without guaranteeing their absence from the auxiliary branch}.

Specifically, we set the meta-network $\bm{\phi}^{\mathcal{B}}$ and the base context $\bm{C}^{\mathcal{B}}$ to be trainable,  while keeping all other parameters fixed. Relying on the established vision--language alignment semantic prior in CLIP, we conduct contrastive learning between the visual feature $\bm{f}$ and the textual embedding $\bm{T}^{\mathcal{B}}$ by
\begin{equation}
\bm{\mathcal{L}}_{\mathrm{pre}} =
- \frac{1}{2N} \sum_{i=1}^N \left(
 \log \frac{e^{u_{ii}}}{\sum_{j=1}^N e^{u_{ij}}}
+ \log \frac{e^{u_{ii}}}{\sum_{j=1}^N e^{u_{ji}}}
\right),
\end{equation}


\smallskip
\noindent\textbf{Second-stage Training.}
At this stage, all trainable modules are jointly optimized under a set of dedicated losses, except that CLIP is adapted via LoRA : 

1) \textit{Geometric Separation Loss.} 
Since the prompts $T^{\mathcal{A}}$ and $T^{\mathcal{B}}$ are designed to play different roles, their induced visual features $f^{\mathcal{A}}$ and $f^{\mathcal{B}}$ should be geometrically separated. We encourage this separation by penalizing the absolute cosine similarity:
\begin{equation}
    \bm{\mathcal{L}}_{\mathrm{gs}}
    = \frac{| \bm{f}^{\mathcal{A}} \cdot \bm{f}^{\mathcal{B}} |}{\|\bm{f}^{\mathcal{A}}\| \cdot \|\bm{f}^{\mathcal{B}}\|}.
\end{equation}

2) \textit{Prompt Diversity Loss.}
Different images likely contain different forgery traces and semantic content, and the prompts should have the ability to depict the variance. This loss is defined as
\begin{equation}
    \bm{\mathcal{L}}_{\mathrm{div}} =
        \mathbb{E}_{i \neq j}\left[
          \mathrm{sim}\!\left(
            \bm{T}^{\mathcal{A}}_i,\bm{T}^{\mathcal{A}}_j\right)
        \right] +
        \mathbb{E}_{i \neq j}\left[
          \mathrm{sim}\!\left(
            \bm{T}^{\mathcal{B}}_i,\bm{T}^{\mathcal{B}}_j\right)
        \right],
\end{equation}
where $\mathrm{sim}$ means cosine similarity.

3) \textit{Cross-modality Alignment Loss.}
A naive alignment strategy would directly maximize the similarity between visual and textual features for both streams. However, this overlooks a key asymmetry in the forgery detection task: \textit{all images contain semantic content, whereas only deepfake images exhibit forgery traces}. Therefore, we align $\bm{f}^{\mathcal{B}}$ with $\bm{T}^{\mathcal{B}}$. But for $\bm{f}^{\mathcal{A}}$, only those from fake images are encouraged to align with $\bm{T}^{\mathcal{A}}$, while those from real images are explicitly pushed away from it. This process can be formulated as
\begin{equation}
\begin{array}{cl}
    \bm{\mathcal{L}}_{\mathrm{align}}
    = & -\mathbb{E}\left[
        \mathrm{sim} \left(\bm{f}^{\mathcal{B}}_i,
        \sigma(\bm{T}^{\mathcal{B}}_i) \right)
      \right] \\[0.1cm]
      & -\mathbb{E}\!\left[
        \mathrm{sim}\!\left(\bm{f}^{\mathcal{A}}_i,
        \sigma(\bm{T}^{\mathcal{A}}_i) \right) \cdot \Pi(y_i=1)  \right] \\[0.1cm]
       & + \mathbb{E}\!\left[
        \mathrm{sim}\!\left(\bm{f}^{\mathcal{A}}_i,
        \sigma(\bm{T}^{\mathcal{A}}_i) \right) \cdot \Pi(y_i=0)  \right], \\
\end{array}
\end{equation}
where $\sigma(\cdot)$ denotes a learnable projection layer that maps textual features into the visual feature space, $y_i$ represents the ground-truth label, with $1$ for fake and $0$ for real, and $\Pi$ denotes the indicator function.



\smallskip\noindent\textbf{Overall Objective.}
Similar to existing methods~\cite{svd,forensics_adapter}, we apply a PEFT strategy such as LoRA~\cite{lora} on CLIP with a supervised contrastive loss as $\bm{\mathcal{L}}_{\mathrm{con}}$, and the standard cross-entropy loss on $\bm{f}^{\mathcal{A}}$ as $\bm{\mathcal{L}}_{\mathrm{cls}}$. The overall objective can be written as
\begin{equation}
    \bm{\mathcal{L}}_{\mathrm{all}}
    = \bm{\mathcal{L}}_{\mathrm{cls}}
    + \lambda_1\bm{\mathcal{L}}_{\mathrm{gs}}
    + \lambda_2\bm{\mathcal{L}}_{\mathrm{div}}
    + \lambda_3\bm{\mathcal{L}}_{\mathrm{align}}
    + \lambda_4\bm{\mathcal{L}}_{\mathrm{con}},
\end{equation}
where $\lambda_1, \ldots, \lambda_4$ are the hyperparameters balancing each term.


\section{Experiments}
\label{sec:exp}

\subsection{Experimental Setup}
\label{ssec:setup}

\noindent\textbf{Datasets and Metrics.}
For the cross-dataset setting, our method is trained on FaceForensics++ (c23)~\cite{ffpp} and tested on the Celeb-DF-v2~\cite{celebdf}, DFD~\cite{dfd}, DFDC~\cite{dfdc}, DFDCP~\cite{dfdcp}, and WDF~\cite{wdf} datasets. For the cross-method setting, our method is trained on the FF++ (c23) and tested on various manipulation methods, including UniFace, BleFace, MobSwap, e4s, FaceDan, FSGAN, InSwap, and SimSwap following~\cite{svd}.
We adopt the Area Under the ROC Curve (AUC) at the video level as the primary evaluation metric, following the standard protocol in deepfake detection literature~\cite{repdfd,udd,forensics_adapter}. 


\smallskip\noindent\textbf{Implementation Details.}
All experiments are conducted on a single NVIDIA GeForce RTX 3090 GPU using CUDA 11.3 and PyTorch 1.11.0. Our method adopts the CLIP-ViT-Large-Patch14 backbone with binary class outputs. Data augmentation is applied with random horizontal flipping, rotation, blur, and brightness/contrast perturbation, following~\cite{svd}. The batch size is set to 24 for both training and testing. We employ the Adam optimizer with a learning rate of $2 \times 10^{-4}$, and a weight decay of $5 \times 10^{-4}$ for regularization. For the parameters in the loss function, we set $\lambda_{1}=0.05$, $\lambda_{2}=0.01$, $\lambda_{3}=0.1$, $\lambda_{4}=0.1$. We adopt a dynamic loss weighting strategy with a warm-up ratio of $10\%$. The architecture of the meta-network is a two-layer MLP with a hidden size of 256. The number of learnable context tokens is set to 16.

\subsection{Results}
\label{ssec:results}

Table~\ref{tab:cross_dataset} reports the cross-dataset and cross-manipulation results. The results of other methods are cited from~\cite{svd},~\cite{uclvm} and~\cite{forensics_adapter}. In cross-dataset evaluation, our method achieves the best AUCs on CDF-v2 and DFD, while slightly underperforming ForAda on DFDC and DFDCP. This may be due to the substantial real-world variations and compression artifacts in these datasets, which hinder prompt learning. In contrast, ForAda benefits from an additional visual adapter. Nevertheless, our lightweight method matches or surpasses other state-of-the-art approaches, demonstrating strong generalization across data distributions.

In cross-manipulation evaluation, our method performs best on every forgery type and achieves an average AUC of 0.976, outperforming ForAda by $2.7\%$. The gain is particularly pronounced on BleFace, reaching approximately $8\%$. These results suggest that separable prompt learning effectively captures both manipulation-specific cues, such as blending artifacts, and generic forgery traces.

\begin{figure}[!t]
\centering
\includegraphics[width=0.9\linewidth]{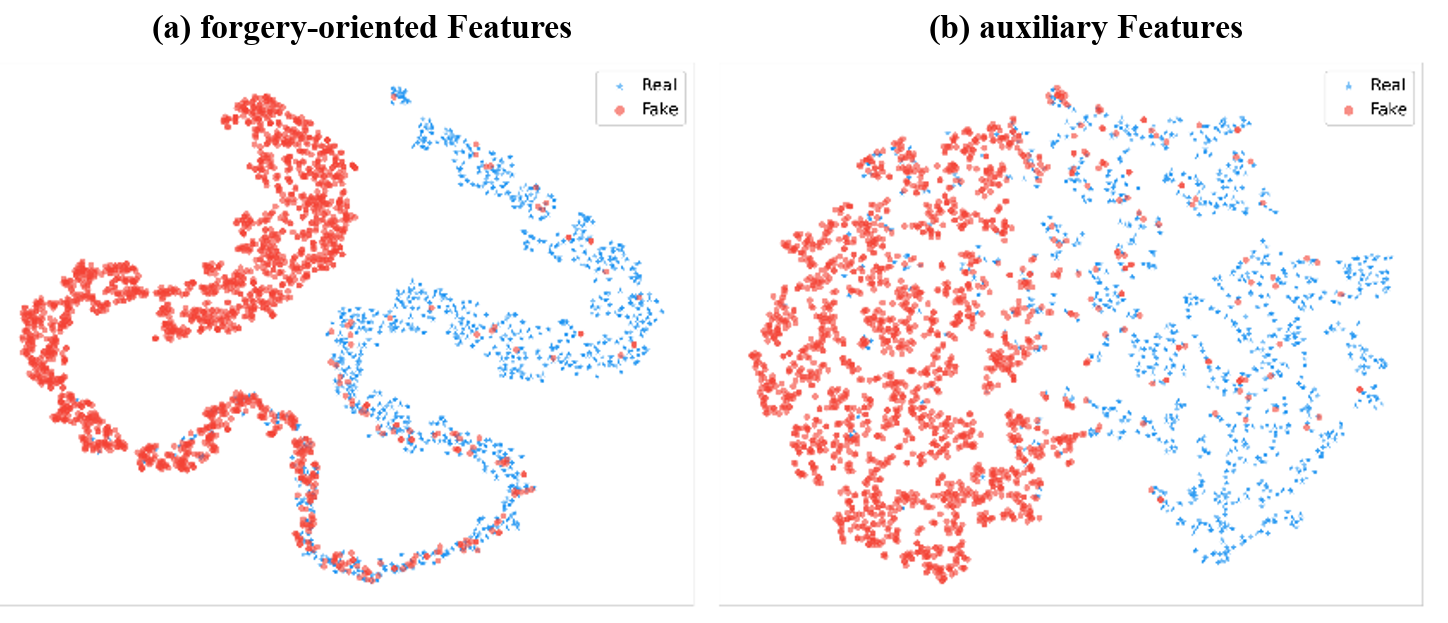}
\vspace{-0.3cm}
\caption{t-SNE visualization of feature distributions (a) $\bm{f}^{\mathcal{A}}$ and (b) $\bm{f}^{\mathcal{B}}$ on FF++ c23. }
\label{fig:p2}
\end{figure}

Figure~\ref{fig:p2} visualizes feature distributions on the FF++ c23, where (a) and (b) represent $\bm{f}^{\mathcal{A}}$ and $\bm{f}^{\mathcal{B}}$ respectively. It can be seen that $\bm{f}^{\mathcal{A}}$ has a better feature separation than  $\bm{f}^{\mathcal{B}}$, demonstrating its effectiveness.

\subsection{Analysis}


\smallskip
\noindent\textbf{Ablations.} \textit{1) Effect of Loss Terms:}
Table~\ref{tab:ablation} progressively adds each loss to the LoRA-based CLIP baseline trained with $\bm{\mathcal{L}}_{\mathrm{cls}}$. Adding $\bm{\mathcal{L}}_{\mathrm{con}}$ improves average performance by approximately $0.6\%$, while $\bm{\mathcal{L}}_{\mathrm{gs}}$ yields further gains. The alignment loss $\bm{\mathcal{L}}_{\mathrm{align}}$ provides an additional average improvement of $0.35\%$, and $\bm{\mathcal{L}}_{\mathrm{div}}$ achieves the best overall results. Together, these losses improve the average AUC by $1.2\%$ over the baseline.
\textit{2) Variants of CLIP Backbones:}
Table~\ref{tab:vlm_comparison} compares three CLIP backbones on CDF-v2, DFD, and DFDCP. Performance improves with model capacity, with ViT-L/14 outperforming ViT-B/16 and ViT-B/32 by $4.6\%$ and $7.5\%$ on average, respectively. These results support the use of ViT-L/14 as the default backbone.

\begin{table}[!t]
\centering
\caption{Ablations across datasets.}
\vspace{-0.3cm}
\label{tab:ablation}
\resizebox{\columnwidth}{!}{%
\begin{tabular}{cccc|cccc}
\toprule
$L_{\rm con}$ & $L_{\rm gs}$ & $L_{\rm align}$ & $L_{\rm div}$ &
CDF-v2 & DFD & DFDC & DFDCP \\
\midrule
$\times$ & $\times$ & $\times$ & $\times$ & 0.942 & 0.965 & 0.845 & 0.903 \\
$\checkmark$ & $\times$ & $\times$ & $\times$ & 0.957 & 0.967 & 0.852 & 0.902 \\
$\checkmark$ & $\checkmark$ & $\times$ & $\times$ & 0.957 & 0.960 & 0.859 & 0.904 \\
$\checkmark$ & $\checkmark$ & $\checkmark$ & $\times$ & 0.959 & 0.970 & 0.863 & 0.902 \\
$\checkmark$ & $\checkmark$ & $\checkmark$ & $\checkmark$ &
\textbf{0.960} & \textbf{0.972} & \textbf{0.866} & \textbf{0.905} \\
\bottomrule
\end{tabular}}
\end{table}

\begin{table}[!t]
\centering
\caption{Effect of using various CLIP Backbones.}
\vspace{-0.3cm}
\small
\label{tab:vlm_comparison}
\setlength{\tabcolsep}{8pt}
\begin{tabular}{l|ccc}
\toprule
VLMs & CDF-v2 & DFD & DFDCP \\
\midrule
CLIP ViT-B/16  & 0.898 & 0.957 &  0.844 \\
CLIP ViT-B/32  & 0.883 & 0.899 &  0.830 \\
CLIP ViT-L/14  & \textbf{0.960} & \textbf{0.972} & \textbf{0.905} \\
\bottomrule
\end{tabular}
\end{table}

\begin{figure}[!t]
\centering
\includegraphics[width=\columnwidth]{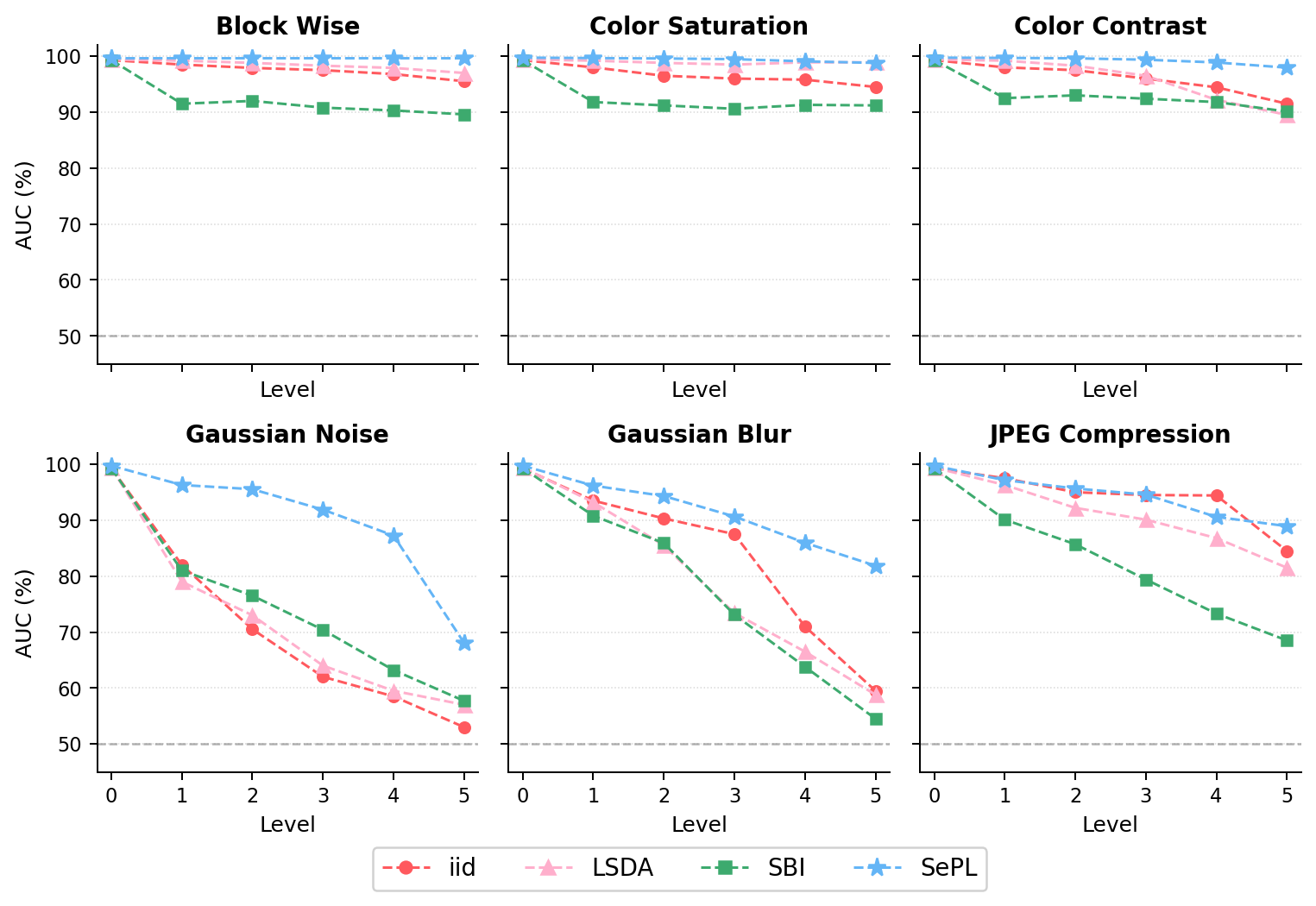}
\vspace{-0.6cm}
\caption{Robustness evaluation under six types of image perturbations at five severity levels.}
\label{fig:r1}
\end{figure}

\smallskip\noindent\textbf{Robustness.}
Following~\cite{forensics_adapter}, we evaluate robustness under six perturbations at five severity levels, including block-wise masking, color saturation, color contrast, Gaussian noise, Gaussian blur, and JPEG compression. As shown in Fig.~\ref{fig:r1}, SePL achieves the highest AUC under most settings and degrades more gracefully than IID~\cite{iid} and SBI~\cite{sbi} as severity increases. These results demonstrate improved robustness to the evaluated image perturbations.

\section{Conclusion}
\label{sec:conclusion}

In this paper, we present SePL, a new CLIP-based face forgery detector enhanced by separable prompt learning. Specifically, SePL introduces two types of learnable prompts, designed to distill forgery knowledge from CLIP features. To achieve this, we design a cross-modality alignment strategy and a set of dedicated loss terms. SePL is extensively evaluated on a wide range of datasets and compared against several recent face forgery detection methods. The superior performance demonstrates the effectiveness of our method in improving generalization for face forgery detection. 

\vfill\pagebreak

\small{
\bibliographystyle{IEEEbib}
\bibliography{refs}
}

\end{document}